\documentclass[letterpaper, 10 pt, conference]{ieeeconf}  

\IEEEoverridecommandlockouts                              

\usepackage{graphicx} 
\graphicspath{{./pdf/}} 
\DeclareGraphicsExtensions{.pdf,jpg}
\usepackage[export]{adjustbox} 
\usepackage[skip=1pt,labelformat=empty]{subcaption} 
\usepackage[cmex10]{amsmath} 
\usepackage[short]{optidef} 
\usepackage[ruled,linesnumbered]{algorithm2e} 
\usepackage{caption} 
\usepackage{bm} 
\usepackage{epsfig} 
\usepackage{times} 
\usepackage{color} 
\usepackage{amssymb}  
\usepackage{setspace}

\newcommand{\mytilde}{\raise.17ex\hbox{$\scriptstyle\mathtt{‌​\sim}$}}
\usepackage{xparse,soul}
\usepackage{multirow}
\usepackage{cite}

\usepackage{titlesec}
\titlespacing{\section}{0pt}{5pt}{1pt}
\titlespacing{\subsection}{0pt}{2pt}{1pt}
\titlespacing{\subsubsection}{10pt}{0pt}{*0}
\titleformat{\subsubsection}[runin]{\slshape}{}{}{}

\title{\LARGE \bf
FBG-Based Variable-Length Estimation for Shape Sensing of Extensible Soft Robotic Manipulators
}

\author{Yiang Lu, Wei Chen, Zhi Chen, Jianshu Zhou, and Yun-hui Liu
\thanks{This work was supported in part by Shenzhen Portion of Shenzhen-Hong Kong Science and Technology Innovation Cooperation Zone under HZQB-KCZYB-20200089, in part of the HK RGC under T42-409/18-R and 14202918, in part by the Hong Kong Centre for Logistics Robotics, in part by the Multi-Scale Medical Robotics Centre, InnoHK, and in part by the VC Fund 4930745 of the CUHK T Stone Robotics Institute. \textit{(Corresponding author: Jianshu Zhou. Email: jianshuzhou@cuhk.edu.hk)}}
\thanks{Y. Lu, W. Chen, Z. Chen, J. Zhou, and Y.-H. Liu are with T Stone Robotics Institute, Department of Mechanical and Automation Engineering, The Chinese University of Hong Kong, Hong Kong.}%
\thanks{J. Zhou and Y.-H. Liu are also with the Hong Kong Center for Logistics Robotics, Hong Kong.}%
}

\begin{document}

\maketitle
\thispagestyle{empty}
\pagestyle{empty}

\begin{abstract}
In this paper, we propose a novel variable-length estimation approach for shape sensing of extensible soft robots utilizing fiber Bragg gratings (FBGs).
Shape reconstruction from FBG sensors has been increasingly developed for soft robots, while the narrow stretching range of FBG fiber makes it difficult to acquire accurate sensing results for extensible robots.
Towards this limitation, we newly introduce an FBG-based length sensor by leveraging a rigid curved channel, through which FBGs are allowed to slide within the robot following its body extension/compression, hence we can search and match the FBGs with specific constant curvature in the fiber to determine the effective length.
From the fusion with the above measurements, a model-free filtering technique is accordingly presented for simultaneous calibration of a variable-length model and temporally continuous length estimation of the robot, enabling its accurate shape sensing using solely FBGs.
The performances of the proposed method have been experimentally evaluated on an extensible soft robot equipped with an FBG fiber in both free and unstructured environments. 
The results concerning dynamic accuracy and robustness of length estimation and shape sensing demonstrate the effectiveness of our approach.
\end{abstract}

\section{INTRODUCTION}

Soft robotic manipulators have received massive attention owing to their inherent compliance and safety endowing their unique advantages in versatile applications, e.g., industrial fields, surgical interventions \cite{Fangeabg5575}, and human-robot interactions \cite{yi2018customizable}.
Among them, extensible soft manipulators composed of elastic chambers or pneumatic actuators/muscles are capable of changing their lengths in wide ranges\cite{mcmahan2005design, cianchetti2012design, cianchetti2013stiff, elena2018novel, chen2019design}, allowing them to achieve more flexible motions, i.e., elongation and shortening, and also accompanied with variable stiffness.
Taking the length variation into consideration, shape reconstruction of extensible manipulators is regarded as more complicated than those counterparts with fixed lengths. 
Therefore, achieving accurate measurements of their variable length and shape are still highly deserved topics, which can provide reliable guidance for precise control and manipulation tasks, especially when performing in unknown and constrained environments with unexpected external forces and perturbations \cite{wang2016visual, takaki2019acoustic}.

Previous efforts have been made for variable length sensing of extensible soft manipulators.
For example, model-based approaches can calculate the shape (length) of soft robot from the kinematics relationship (e.g., constant curvature model), while they require the sensing data in the joint space, such as cable length from encoder and pneumatic pressure \cite{chou1996measurement, webster2010design, chen2019design, lai2021constrained}.
The performance of model-based methods, as well as their length estimation accuracy, are significantly associated with the accuracy of soft continuum robot modeling. 
However, the accurate modeling of soft manipulator composed of nonlinear materials, especially when undergoing large deflections or interacting in unstructured environments, is another unsolved challenging problem.
To cope with this, Wang \textit{et al.} \cite{wang2016visual} proposed an adaptive controller for eye-in-hand visual servoing of soft robot in constrained scenarios, where the adaptation algorithm can online estimate the robot length using vision feedback and constant curvature model \cite{wang2013visual}.
Alternatively, novel sensor-based approaches are also proved effective for variable-length sensing in the field of soft robotics, among which stretchable structures and materials have been utilized for design and fabrication of length sensors.
Nakamoto \textit{et al.} \cite{nakamoto2015design} designed a stretchable strain sensor to acquire the length for a pneumatic artificial muscle (PAM), where an inverting amplifier circuit was designed for capacitance measurement with fast response characteristics.
Park and Wood \cite{park2013smart} measured PAM length mapped from the resistance of a liquid conductor, which was embedded in the PAM helical microchannel.
Conductive yarns were wrapped around the soft robot to determine its elongation from inductance measurement \cite{felt2015contraction}.
For these sensors integrating conductive materials, the limited resolution and linearity may hamper their practical extensions to the robot capable of large deformation \cite{takaki2019acoustic}.
Besides, they can only obtain the elongation in the length,  while detecting deformations in other directions (e.g, deflections) require extra sensing information.
To \textit{et al.} \cite{to2015highly} presented a soft optical sensor based on light intensity modulation to measure both stretching and bending deformations.
A length sensor based on acoustic resonance was introduced for a soft continuum arm, which is composed of a speaker and a microphone with a frequency characteristic model \cite{takaki2019acoustic}.

Recently, fiber Bragg grating (FBG) sensors have been widely applied for shape reconstruction of continuum robots and soft robots \cite{roesthuis2013three,wang2016shape,shi2016shape,denasi2018observer,khan2019multi,rahman2019modular,alambeigi2019scade,wang2020eye,monet2020high,sefati2020data,chitalia2020towards,sefati2021dexterous,lu2021robust} due to their various advantages, such as small size, high biocompatibility, and high-frequency streaming rate without requiring line-of-sight.
Wang \textit{et al.} \cite{wang2016shape} introduced a shape sensing approach for a soft robot equipped with an FBG sensor network to measure the robot curvature and torsion.
Monet \textit{et al.} \cite{monet2020high} employed optical frequency domain reflectometry (OFDR) technique for shape reconstruction of a continuum robot, which enables higher spatial resolution and accurate sensing results.
Alambeigi \textit{et al.} \cite{alambeigi2019scade} proposed a model-independent Kalman filter (KF) by fusing FBG data with eye-to-hand vision, in order to simultaneously estimate the deformation model of a continuum manipulator as well as its tip position.
A learning-based motion estimator was designed by combining 2D imaging with sparse FBG strains for a soft manipulator \cite{wang2020eye}, which alleviates the need for precise alignment of FBG fiber on the robot.
A model-based filtering algorithm for shape sensing using multi-core FBGs was recently reported to be accurate and robust against the sensory noises and unexpected perturbations \cite{lu2021robust}.
However, owing to the narrow extension range of the optical fiber, the existing approaches that only adopt FBG sensors cannot be directly applied for extensible soft robot with variable length to prevent the fiber broken.
It is possible that the state-of-the-art works using multiple modalities \cite{alambeigi2019scade,wang2020eye} could handle such conditions,  but stretching deformation of FBG fiber is still limited in a small range and the feasible solution has not been developed.

Targeting these limitations, we propose a novel FBG-based variable-length estimation approach for shape sensing of extensible soft manipulators. To the best of our knowledge, this is the first work for simultaneous effective length sensing and shape reconstruction by using solely FBG sensors. Our main contributions can be summarized as follows:
\begin{enumerate}
    \item To enable length sensing from FBGs and avoid the fiber broken, we newly design an FBG-based length sensor for extensible soft robot, which is composed of a rigid channel with specific constant curvature and is easy to be deployed on the robot. FBGs in the fiber can slide through the channel with extension and compression of the robot, hence the gratings with the specific curvature can be matched and located to measure the robot length.
    \item By leveraging the intermittent but accurate results from the proposed length sensor as measurements, a model-free Kalman filter (KF) is introduced to not only calibrate a variable-length model of the soft robot, but also simultaneously estimate the effective length with guaranteed temporal continuity, which can be further utilized for accurate shape sensing.
    \item Our method is generic and independent of extra sensors (e.g., cameras) while it ensures high accuracy and robustness in dynamic conditions and unstructured environments as compared with the model-based method, thus proving its superiority and application feasibility for various soft manipulators with variable length.
\end{enumerate}

The rest of this paper is organized as follows:
Section II introduces the design of an FBG-based length sensor.
Section III details a variable-length model and our proposed model-free filtering technique.
Experiments and results are evaluated and discussed in Section IV.
Conclusions and future works are presented in Section V.

\section{Design of FBG-Based Length Sensor}
The proposed approach for variable-length estimation and shape sensing of extensible soft robots is mainly comprised of two novel components as shown in Fig. \ref{fig:diagram}: a) an FBG-based length sensor and b) a model-free filtering algorithm for variable-length estimation.
In this section, we first review the previous study on curvature/twist estimations using FBG sensors, the results of which can be utilized for our length sensor to be subsequently introduced.

\subsection{FBG Curvature/Twist Estimations for Shape Sensing} 
In the previous shape sensing method using multi-core FBG sensors, a model-based filtering technique was introduced for iterative curvature/twist estimations and accurate shape reconstruction of continuum manipulators, which can deal with noisy signals, accumulated error with sensing length, and external disturbances \cite{lu2021robust}. 
For shape sensing, the curvatures $\kappa_s$ and twists $\tau_s, s \in \{1, 2, ..., M\}$ at $M$ FBG-sections on the fiber are accordingly estimated from the previous method.

In this paper, the total number of FBG-sections $M$ should be determined by the FBG spatial resolution $\lambda$ together with the maximum length $l_{max}$ of the extensible soft robot and the arc length $l_{c}$ of the channel in our length sensor, satisfying $l_{max} + l_{c} = M \lambda$.
The entire shape with variable sensing length represented by the pose $\bm{T}_{s} \in SE(3)$, $s \in \{\mu, \mu + 1,  ..., M\}$, i.e., the position and orientation of each FBG-section in a global frame, can be consequently reconstructed using a series of homogeneous transformations ${}^{s-1}\bm{T}_{s} \in SE(3)$, $s \in \{\mu + 1, \mu + 2, ..., M\}$, from the origin (starting index $\mu$) to the distal endpoint given by
\begin{align}
    \label{eq:transform}
    & {}^{s-1} \bm{T}_{s}=
    \begin{bmatrix}
        {}^{s-1} \bm{R}_{s} & \bm{p}_{s} \\
        \bm{0}_{1 \times 3} & 1
    \end{bmatrix} \\ 
    & \quad \; \; \bm{p}_{s} = 
    \begin{bmatrix}
        \rho_s - \rho_s \mathrm{cos}(\theta_s) & 0 & \rho_s \mathrm{sin}(\theta_s)
    \end{bmatrix}^{\intercal} \\ 
    & {}^{s-1} \bm{R}_{s} = \mathrm{Rot}(\widehat{z},\tau_s)  \mathrm{Rot}(\widehat{y},\theta_s) 
\end{align}
where $\bm{p}_{s} \in \mathbb{R}^{3}$ and ${}^{s-1} \bm{R}_{s} \in SO(3)$, $s \in \{\mu + 1, \mu + 2, ..., M\}$, are the relative translations and rotations of each FBG-section in the local frame, $\rho_s =  1/\kappa_s$ is the radius of curvature, and $\theta_s = \kappa_s \Delta s$ is the bending angle. $\mathrm{Rot}(\widehat{z},\tau_s)$ and $\mathrm{Rot}(\widehat{y},\theta_s) \in SO(3)$ are the rotation operations by $\tau_s$ and $\theta_s$ about $\widehat{z}$ and $\widehat{y}$ coordinates of the local frame, respectively.
Note that unlike the previous methods with fixed sensing length, here the starting index $\mu$ for shape reconstruction depends on the effective robot length and it can be obtained from the following FBG-based length sensor.

\begin{figure*}[!t]
  \centering
  \setlength{\abovecaptionskip}{0pt}
  \setlength{\belowcaptionskip}{-5pt} 
  \includegraphics[width=0.95\linewidth]{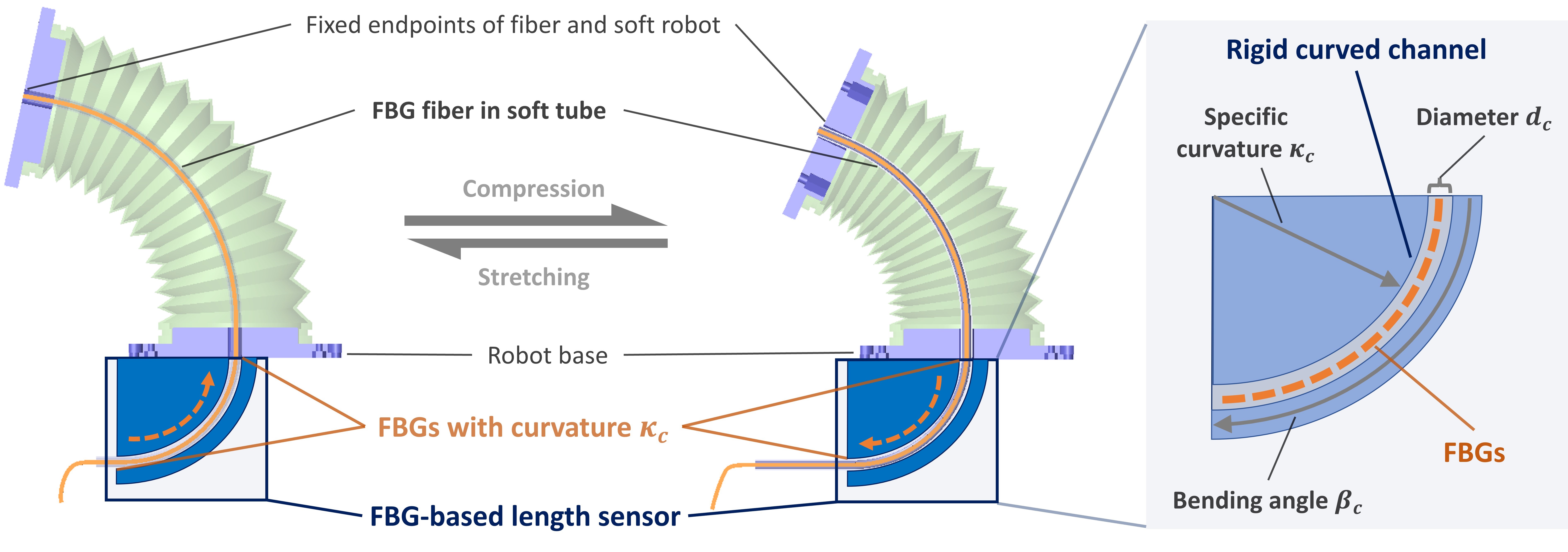}
  \caption{Working principle of the proposed FBG-based length sensor that is attached at the robot base. The distal endpoints of FBG fiber and robot are fixed, while the fiber is not glued at the robot base enabling FBGs to slide through the rigid curved channel following the robot elongation/compression. The consecutive FBGs in the channel with specific constant curvature $\kappa_c$ can be matched and their indexes $\mathcal{C}$ can be located to determine the effective lengths of both robot and FBG fiber.}
  \label{fig:principle}
\end{figure*}

\subsection{FBG-Based Length Sensor with Constant Curvature} 
Employing the curvatures $\kappa_s$, $s \in \{1, 2, ..., M\}$ at overall $M$ FBG-sections on the fiber, we design a novel FBG-based length sensor to measure the length variation of the extensible soft manipulator, the working principle of which is demonstrated in Fig. \ref{fig:principle}.
The proposed sensor is mainly composed of a rigid channel with a specific and constant curvature $\kappa_c$, which is installed at the robot base, and the entire sensor can be fabricated by 3-D printing thus making it easy to be deployed.
Unlike the previous shape sensing approach \cite{lu2021robust}, in which two endpoints of the FBG fiber were fixed with the distal point and the base of the continuum manipulator, respectively, when using the proposed sensor, we only attach their distal endpoints, while another fiber endpoint is free to slide through the rigid channel, as a result preventing the fiber broken and endowing length variation sensing.
To keep the FBG fiber sliding smoothly in the rigid channel, the channel structure parameters, such as $\kappa_c$ and bending angle $\beta_c$ as shown in Fig. \ref{fig:principle}, are chosen from some preliminary tests, the values of which will be detailed in the experiment setup.
Additionally, its diameter $d_c$ is dependent on that of FBG fiber and should also guarantee the fiber's smooth movement. 

When the soft robot stretches or compresses, all FBGs can slide within the robot led by the fiber sliding through the channel,
and we can search the consecutive $L_c$ FBG-sections in the channel that are supposed to be bent with the specific curvature, and acquire their indexes $\mathcal{C} = \{\mu - L_c, \mu - L_c + 1, ..., \mu\}$ by
\begin{align}
    \label{eq:min}
    \underset{\mu}{\text{argmin}} \;
    \left\| \sum_{i = \mu - L_c}^{\mu}
    \kappa_i - \kappa_c L_c \right\|
\end{align}
where minimizing the above objective function represents matching the consecutive $L_c$ FBGs, whose sum of curvature should conform to the sum of specific curvature $\kappa_c L_c$ with acceptable estimation errors.
A simple detection algorithm can be employed to exclude the wrong index far from the result in the last iteration, which may occur when the soft robot deflects to a similar curvature, and such a deviation can be alternatively resolved by the following model-free filtering method.
Consequently, the effective sensing length $l_{e}$ is obtained from the above result and given by 
\begin{align}
    \label{eq:length}
    l_{e} = l_{max} - \lambda \mu
\end{align}
and the shape of the robot can be reconstructed by using the curvatures $\kappa_s$ and twists $\tau_s, s \in \{\mu, \mu + 1, ..., M\}$ from (\ref{eq:transform}).
Note that the length sensing result $l_{e}$ is accurate which will be proved in Section IV, but it could be intermittent due to spatial resolution $\lambda$ of the FBG fiber.
Therefore, by leveraging $l_{e}$ as measurement, a model-independent filter is deployed for simultaneous calibration of a variable-length model and effective length estimation with guaranteed temporal continuity.

\begin{figure*}[!t]
  \centering
  \setlength{\abovecaptionskip}{0pt}
  \setlength{\belowcaptionskip}{-5pt} 
  \includegraphics[width=0.85\linewidth]{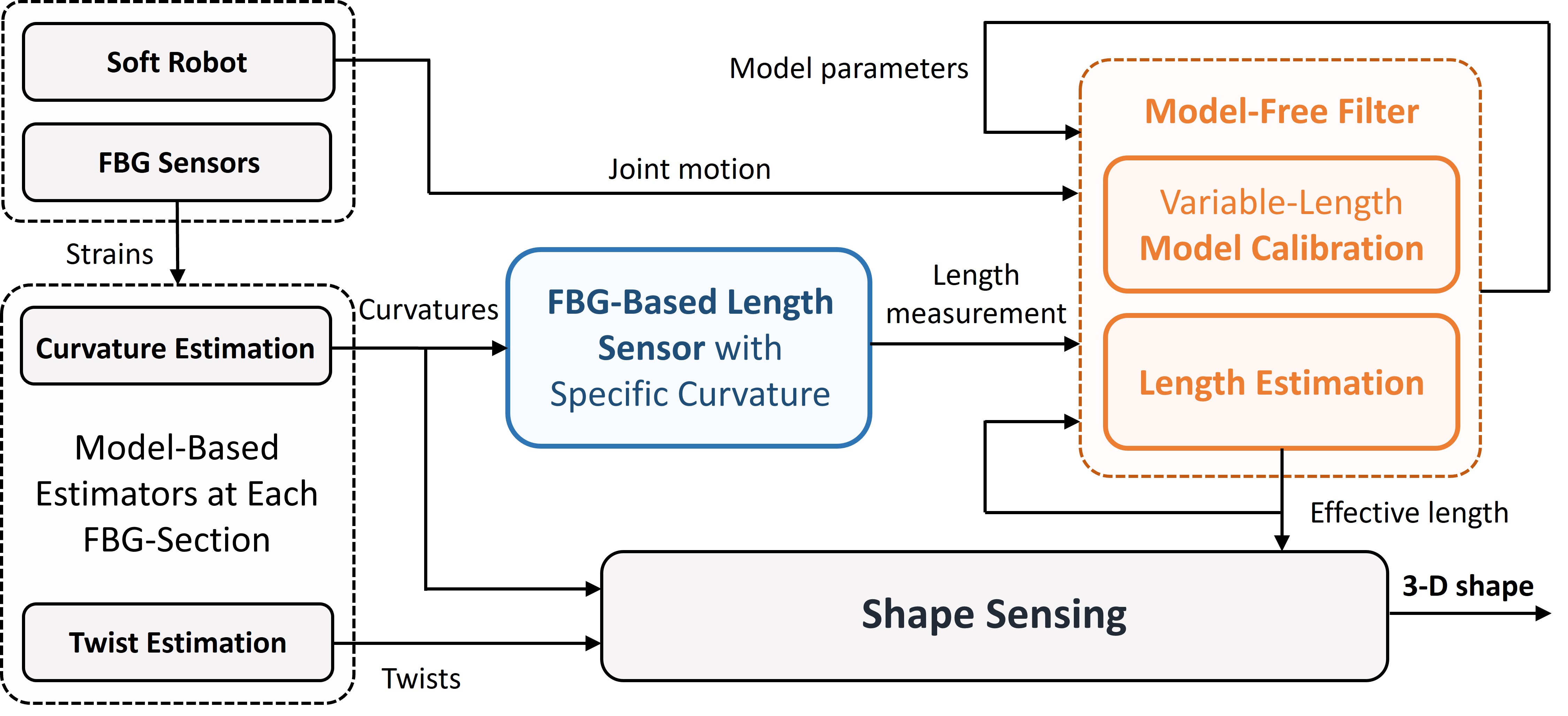}
  \caption{Block diagram of the proposed FBG-based framework for shape sensing with variable effective length, where the length sensor with specific constant curvature can acquire accurate length measurement from the estimated FBG curvatures, and the model-free filter can continuously estimate the effective length of soft robot for accurate shape sensing results.}
  \label{fig:diagram}
\end{figure*}

\section{Model-Free Filter for Variable-Length Model Calibration and Shape Sensing}

In this section, a novel model-free filtering technique is proposed by using the above length measurement as shown in Fig. \ref{fig:diagram}, which can calibrate a variable-length model and simultaneously estimate the continuous effective length for accurate shape sensing.

\subsection{Variable-Length Modeling} 
Define the effective length $l \in \mathbb{R}$ of the extensible soft robot, and a vector of its actuation positions $\bm{q} = \begin{bmatrix} q_1 & q_1 & \cdots & q_n \end{bmatrix}^{\intercal} \in \mathbb{R}^{n}$.
We formulate the kinematics of the soft robot mapping the actuators $\bm{q}$ to the length by a smooth function as $l = l(\bm{q}): \mathbb{R}^{n} \mapsto \mathbb{R}$, and its differential kinematics as the variable-length model \cite{webster2010design,chen2021tele} is 
\begin{align}
    \label{eq:kinematics}
    \dot{l} = 
    \bm{\mathcal{J}} (\bm{q}) \dot{\bm{q}}
\end{align}
where $\bm{\mathcal{J}} (\bm{q}) \in \mathbb{R}^{1 \times n}$ denotes the length Jacobian matrix involving the factors related to variable length, e.g., transmission ratio from the joint positions to the robot length.

To acquire the length of extensible soft robot using such a model-based method, a priori identification of the robot model is required, which is time-varying and may become inaccurate due to inherent nonlinear properties as well as unknown external disturbances.
Although with these limitations, the model-based approach has well continuity of prediction in the temporal domain, and this advantage can be taken into a filtering algorithm in a fusion with the intermittent but accurate length measurement from the proposed FBG-based length sensor.
Observe that we can make the model parameters in $\bm{\mathcal{J}} (\bm{q})$ of the variable-length model appear linearly as
\begin{align}
    \label{eq:1}
    \dot{l} = 
    \dot{\bm{q}}^{\intercal}
    \bm{\mathcal{J}}^{\intercal} (\bm{q}) 
\end{align}
which allows us to simultaneously update inaccurate model parameters in $\bm{\mathcal{J}} (\bm{q})$ and estimate the effective length $l$.

We define the state vector $\bm{x}_t \in \mathbb{R}^{2n+1}$ at time instant $t$ by stacking the length $l_t$, model parameters (i.e., the transpose of length Jacobian $\bm{\mathcal{J}}_t (\bm{q})$, and the transpose of its evolution term $\bm{\delta}_t \in \mathbb{R}^{1 \times n}$ given by
\begin{align}
    \label{eq:states}
    \bm{x}_{t}
    =
    \begin{bmatrix}
    l_t & \bm{\mathcal{J}}_t (\bm{q}) & \bm{\delta}_t
    \end{bmatrix}^{\intercal}
\end{align}
where $\bm{\delta}_t$ is time-varying and unknown that describes the slowly changing of $\bm{\mathcal{J}}_t (\bm{q})$ \cite{alambeigi2019scade}. The state transition model of the system is formulated as
\begin{align}
    \label{eq:process}
    & \bm{x}_{t+1} = 
    \bm{\mathcal{A}} \; \bm{x}_{t} + \bm{w}_{t}
\end{align}
where $\bm{w}_t \sim (\bm{0}, \bm{Q}) \in \mathbb{R}^{2n+1}$ is the process noise assumed to be zero mean multivariate Gaussian noise with covariance $\bm{Q} \in \mathbb{R}^{(2n+1) \times (2n+1)}$.
The state transition matrix $\bm{\mathcal{A}} \in \mathbb{R}^{(2n+1) \times (2n+1)}$ in (\ref{eq:process}) is accordingly derived by
\begin{align}
	\bm{\mathcal{A}} =
    \begin{bmatrix}
		1
		& \dot{\bm{q}}^{\intercal}
		& \bm{0}_{3 \times 3} \\
		\bm{0}_{3 \times 3} 
		& \bm{I}_{3 \times 3}   
		& \Delta t \cdot \bm{I}_{3 \times 3} \\
		\bm{0}_{3 \times 3} 
		& \bm{0}_{3 \times 3} & \bm{I}_{3 \times 3}
	\end{bmatrix}
\end{align}
where $\bm{I}_{3 \times 3} \in \mathbb{R}^{3 \times 3}$ is an identity matrix, $\bm{0}_{3 \times 3} \in \mathbb{R}^{3 \times 3}$ is a zero matrix, and $\Delta t$ is the sampling time.

At time instant $t$, by utilizing the length sensing result $l_e$ from the proposed length sensor (\ref{eq:length}) as the measurement $z_t = l_e \in \mathbb{R}$, we derive the measurement model as 
\begin{align}
    \label{eq:measurement}
    z_t = 
    \bm{\mathcal{H}} \; \bm{x}_t + v_t
\end{align}
where $\bm{\mathcal{H}} \in \mathbb{R}^{1 \times (2n+1)}$ is the measurement matrix given by
\begin{align}
	\bm{\mathcal{H}} =
    \begin{bmatrix}
		1
		& \bm{0}_{1 \times 2n} 
	\end{bmatrix}
\end{align}
and $\bm{0}_{1 \times 2n} \in \mathbb{R}^{1 \times 2n}$ is a zero row vector.
In (\ref{eq:measurement}), $v_t \sim (0, \varUpsilon) \in \mathbb{R}$ is the measurement noise assumed to be zero mean Gaussian noise with variance $\varUpsilon \in \mathbb{R}$.

\subsection{Model-Free Filter for Effective Length Estimation}
Inspired by a filtering algorithm for deformation model calibration of a continuum manipulator and estimation of its endpoint position \cite{alambeigi2019scade}, we propose a model-free Kalman filter (KF) by leveraging the length measurement from our FBG-based sensor and taking advantage of time continuity of the variable-length model.
Therefore, our approach can iteratively estimate the effective length together with variable-length model parameters, and simultaneously reconstruct the accurate shape of extensible soft robot.
Since the length sensing result from our FBG-based length sensor is intermittent while accurate, the proposed filtering algorithm can be performed and switched between two modes based on different sensing conditions, including the first mode at the time instant $t$ when the length sensor outputs a different starting index $\mu_{t}$ from the last iteration $\mu_{t-1}$, which means that the soft robot stretches or compresses in the length over an FBG spatial resolution $\lambda$, and the second mode when the measurement is same as the last one, i.e., $\mu_{t} = \mu_{t-1}$.

\begin{figure*}[htbp]
  \centering
  \setlength{\abovecaptionskip}{0pt}
  \setlength{\belowcaptionskip}{-5pt} 
  \includegraphics[width=0.98\linewidth]{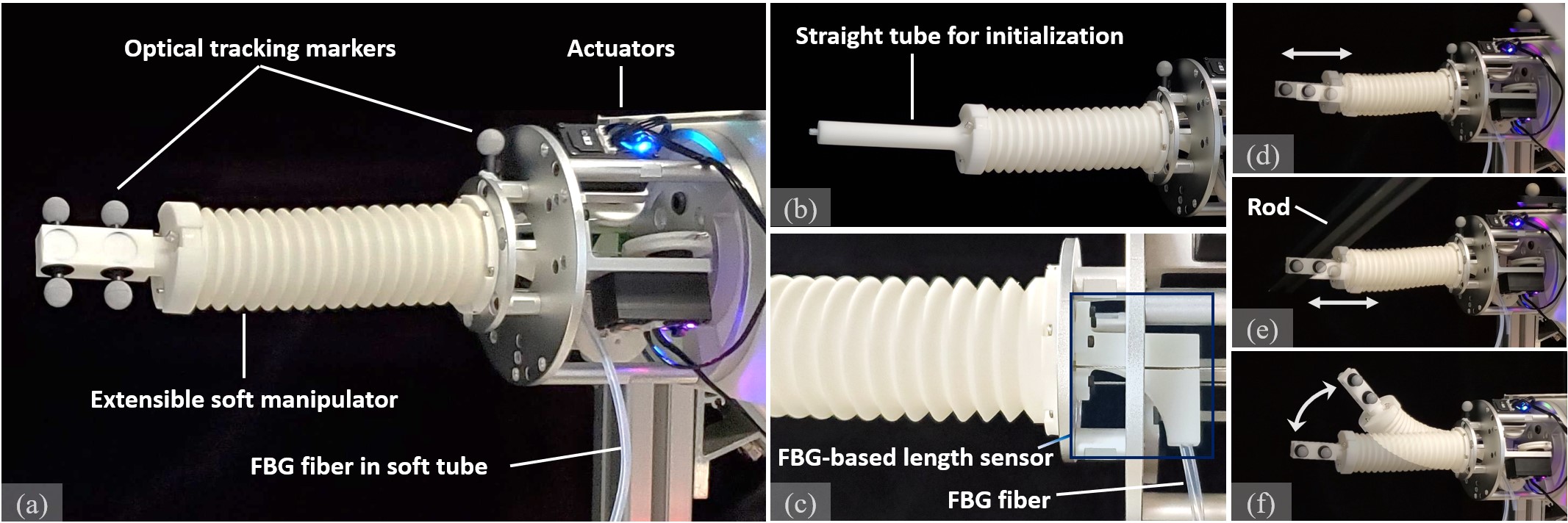}
  \caption{Experimental platform: (a) an extensible soft manipulator equipped with an FBG fiber for shape sensing, and optical tracking markers attached at the robot's distal endpoint and base to acquire the shape ground truth, (b) a straight rigid tube for initialization, and (c) an FBG-based length sensor. Experimental tasks: (d) \texttt{Task 1} length estimation in free space, (e) \texttt{Task 2} length estimation in unstructured environment, and (f) \texttt{Task 3} shape sensing with variable length.}
  \label{fig:expsetup}
\end{figure*}

In the first mode when the length starting index $\mu_{t}$ from the FBG-based sensor is different from the last measurement $\mu_{t-1}$, we first predict the state vector $\widehat{\bm{x}}_{t|t-1}$ and the state covariance matrix $\bm{P}_{t|t-1} \in \mathbb{R}^{(2n+1) \times (2n+1)}$ as
\begin{align}
    \label{eq:predicted1}
    & \widehat{\bm{x}}_{t|t-1} =
    \bm{\mathcal{A}} \; \widehat{\bm{x}}_{t-1|t-1}  \\
    \label{eq:predicted2}
    & \bm{P}_{t|t-1} =
    \bm{\mathcal{A}} \; 
    \bm{P}_{t-1|t-1} 
    \bm{\mathcal{A}}^{\intercal} + \bm{Q}
\end{align}
and then we can leverage the new length measurement $z_t = l_{e,t}$ and update the state vector $\widehat{\bm{x}}_{t|t}$ and its covariance matrix $\bm{P}_{t|t}$ following a standard KF given by
\begin{align}
    \label{eq:kalmangain}
    & \bm{K}_t=
    \bm{P}_{t|t-1} \bm{\mathcal{H}}_t^{\intercal} (\bm{\mathcal{H}}_t \bm{P}_{t|t-1} \bm{\mathcal{H}}_t^{\intercal} + \varUpsilon)^{-1} \\
    \label{eq:update1}
     & \widehat{\bm{x}}_{t|t} =
    \widehat{\bm{x}}_{t|t-1} 
    + \bm{K}_t (z_t - \bm{\mathcal{H}} \; \widehat{\bm{x}}_{t|t-1}) \\
    \label{eq:update2}
    & \bm{P}_{t|t} =
    (\bm{I}_{(2n+1) \times (2n+1)} - \bm{K}_t \bm{\mathcal{H}}_t) \bm{P}_{t|t-1}
\end{align}
where $ \bm{K}_t \in \mathbb{R}^{(2n+1) \times 1}$ denotes the Kalman gain and $\bm{I}_{(2n+1) \times (2n+1)} \in \mathbb{R}^{(2n+1) \times (2n+1)}$ is an identity matrix.

In the second mode when the measured starting index $\mu_t$ is same as the last one $\mu_{t-1}$ (i.e., $\mu_t = \mu_{t-1}$), we can only propagate the state estimate $\widehat{\bm{x}}_{t|t}$ and its covariance $\bm{P}_{t|t} \in \mathbb{R}^{(2n+1) \times (2n+1)}$ by following (\ref{eq:predicted1}) and (\ref{eq:predicted2}) as
\begin{align}
    \label{eq:predicted1_2}
    & \widehat{\bm{x}}_{t|t} =
    \bm{\mathcal{A}} \; \widehat{\bm{x}}_{t-1|t-1}  \\
    \label{eq:predicted2_2}
    & \bm{P}_{t|t} =
    \bm{\mathcal{A}} \; 
    \bm{P}_{t-1|t-1} 
    \bm{\mathcal{A}}^{\intercal} + \bm{Q}
\end{align}
where the length measurement $l_{e,t}$ from our FBG-based sensor together with the update procedures (\ref{eq:update1}) and (\ref{eq:update2}) in the first mode are alleviated, and only the calibrated variable-length model (\ref{eq:kinematics}) is adopted to for effective length prediction, which can guarantee the temporal continuity of the estimation results.

The effective length $l_t$ as well as model parameters in $\bm{\mathcal{J}}_t (\bm{q})$ can be obtained from the estimated state $\widehat{\bm{x}}_{t|t}$ by using either the first mode or the second one at each time instant. 
The starting index $\mu$ is accordingly calculated as the origin to acquire shape from (\ref{eq:transform}) utilizing the estimated curvatures $\kappa_s$ and twists $\tau_s, s \in \{\mu, \mu + 1, ..., M\}$.
The implementation of the proposed framework for 3D shape sensing with variable-length estimation is presented in Algorithm 1.

\let\oldnl\nl
\newcommand{\nonl}{\renewcommand{\nl}{\let\nl\oldnl}}

\begin{algorithm}[!ht]
  \DontPrintSemicolon
  \nonl\textbf{Initialization:}\;
  $ M\longleftarrow$ total number of FBG-sections\;
  $ t \longleftarrow$ time instant $t = 0$\;
  $\bm{\mathcal{J}} (\bm{q}) \longleftarrow$ initial length Jacobian \;
  $ \widehat{\bm{x}}_{t|t} \longleftarrow$ initial state from FBG-based length sensor (\ref{eq:length})\;
  $ \bm{P}_{t|t}, \bm{Q}, \varUpsilon \longleftarrow$ covariance of $\widehat{\bm{x}}_{t|t}$, $\bm{w}_t$, and $v_t$\;
  $Start\_sensing \longleftarrow Start\_sensing =  \mathrm{True}$\;
  \nonl\textbf{Variable-length estimation and shape sensing:}\;
  \While{$Start\_sensing$}{
    $t \longleftarrow t+1$\;
    \nonl Model-based curvature/twist filter  \cite{lu2021robust}: \;
    $\qquad \kappa_s, \tau_s \longleftarrow$ estimated curvatures and twists \;
    \nonl\textbf{FBG-based length sensor:}\;
    $\qquad l_{e,t}, \mu_t \longleftarrow$ length measurement and index (\ref{eq:length}) \;
    \nonl\textbf{Model-free variable-length estimation:}\;
    \eIf{$\mu_t \neq \mu_{t-1}$}{
        $\widehat{\bm{x}}_{t|t-1}, \bm{P}_{t|t-1} \longleftarrow$ predicted state (\ref{eq:predicted1}) and covariance (\ref{eq:predicted2}) \;
        $\bm{K}_{t}, \widehat{\bm{x}}_{t|t}, \bm{P}_{t|t} \longleftarrow$ 
        Kalman gain (\ref{eq:kalmangain}), updated state (\ref{eq:update1}), and covariance (\ref{eq:update2})\;
    }{
        $\widehat{\bm{x}}_{t|t}, \bm{P}_{t|t} \longleftarrow$ predicted state (\ref{eq:predicted1_2}) and covariance (\ref{eq:predicted2_2})\;
    }
    $l_t \longleftarrow$ estimated effective length from $\widehat{\bm{x}}_{t|t}$\;
    \nonl\textbf{Shape reconstruction} (\ref{eq:transform})
  }
  \caption{Length Estimation and Shape Sensing}
\end{algorithm}

\begin{figure*}[ht]
  \centering
  \setlength{\abovecaptionskip}{-1pt}
  \setlength{\belowcaptionskip}{-7pt} 
  \includegraphics[width=\linewidth]{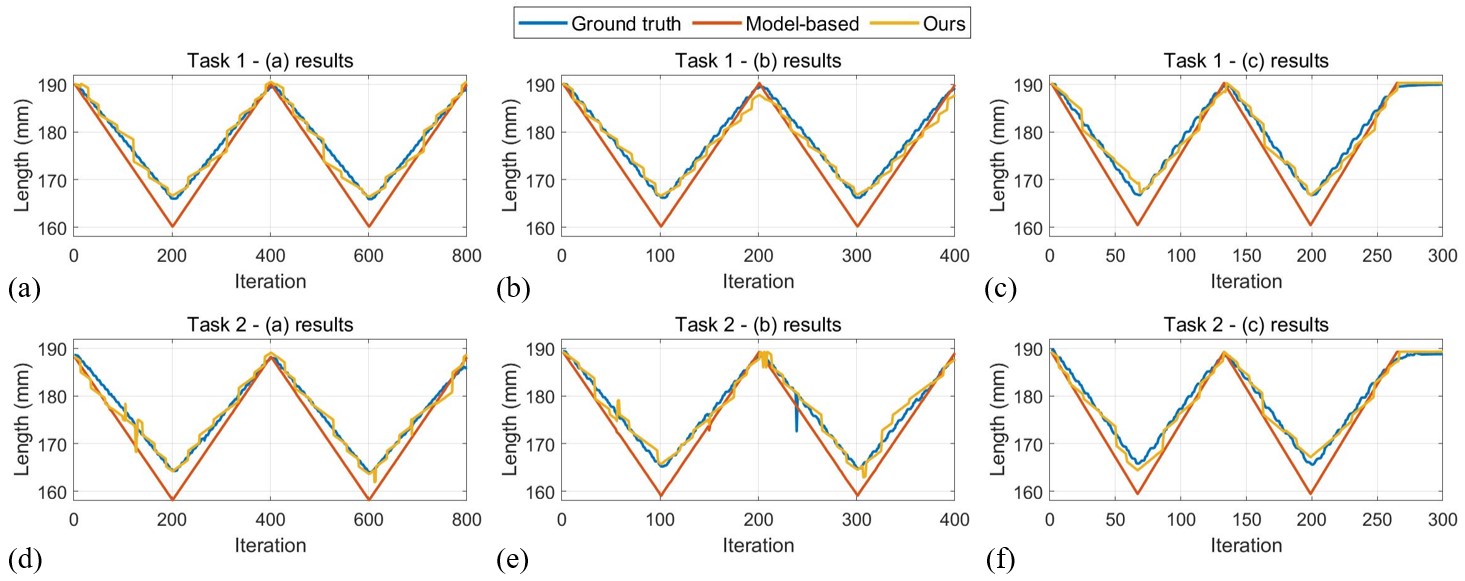}
  \caption{Variable effective length versus time during six trails of (a) $\sim$ (c) \texttt{Task 1} and (d) $\sim$ (f) \texttt{Task 2}.}
  \label{fig:expt1}
\end{figure*}
\begin{figure*}[ht]
  \centering
  \setlength{\abovecaptionskip}{-3pt}
  \setlength{\belowcaptionskip}{-8pt} 
  \includegraphics[width=\linewidth]{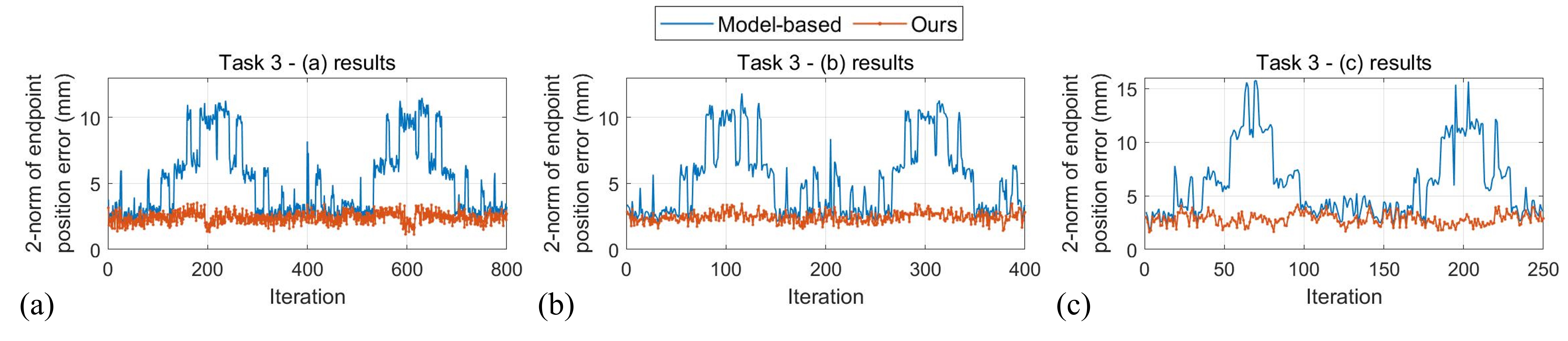}
  \caption{2-norms of distal endpoint position errors versus time during three trails of \texttt{Task 3}-(a), -(b), and -(c).}
  \label{fig:expt3}
\end{figure*}

\section{Experiments}
To evaluate the performance of our approach on an extensible soft manipulator, we compared it with a model-based method, which approximates the robot length from the variable-length model (\ref{eq:kinematics}) as introduced in Section II.
In this section, the experimental setup is firstly detailed, and the results are accordingly presented and discussed.

\subsection{Setup}
The platform for experimental validations of the proposed method consists of an extensible soft manipulator, an FBG-based length sensor, an FBG sensing system, and an optical tracking system as shown in Fig. \ref{fig:expsetup}, the setup of which is extended from that previously reported \cite{lu2021robust}.
We integrated a soft continuum manipulator with variable length for experiments as shown in Fig. \ref{fig:expsetup} (a), which has an outer diameter of 45 mm, a maximum length of 150 mm, and a central soft channel embedded with a multi-core FBG fiber.
The manipulator is composed of the inner and outer backbones fabricated from 3-D printed soft materials (CR-TPU) endowing the manipulator omni-directional deflections driven by three cables and actuated by servo motors (Hiwonder LX-15D) \cite{zhou2020adaptive,chen2021tele}.
Five optical tracking markers in total were attached at the robot's distal endpoint and base, respectively, and four optical trackers (Prime 13, OptiTrack) were used to obtain the position ground truth of these markers, which were calibrated via the motion capture software (Motive, OptiTrack) and the 3D calibration error was 0.062 mm.

To calculate the shape of the soft manipulator, we equipped it with a four-core all grating FBG fiber (FBGS International), whose spatial resolution was $\lambda = 3.3 \; \mathrm{mm}$ and thus 45 FBG-sections (i.e., $M = 45$) were utilized, and their strain signals were simultaneously decoded from an interrogator (RTS125+, Sensuron) with 30 Hz acquisition frequency.
We accordingly estimated the curvatures and twists of these FBG-sections by using the previous model-based filtering algorithm \cite{lu2021robust}, which can be used for effective length sensing and shape reconstruction.
The proposed FBG-based length sensor is demonstrated in Fig. \ref{fig:expsetup} (c), and its structure parameters were selected by preliminary tests as $\kappa_c = 1/30 \; \mathrm{mm}^{-1}$, $\beta_c = 60 ^{\circ}$, which ensure smooth sliding of the FBG fiber through the channel.
Consequently, the implementation of the overall algorithm has 20 Hz sensing frequency comparable with state-of-the-art works \cite{alambeigi2019scade, sefati2020data}.

In the initialization phase before starting our sensing framework we formed the optical fiber in a straight tube as shown in Fig. \ref{fig:expsetup} (b) and recorded the zero strains of FBGs.
The covariance $\bm{R}$ and $\varUpsilon$ for the proposed model-free filtering algorithm were tuned based on the preliminary experiments with desired performance.

\subsection{Results and Discussion}

\begin{table*}[]
\caption{Experimental results: \textbf{absolute value of length error (mm)} for \texttt{Task 1} and \texttt{Task 2}}
\small
\label{table:expt1}
\centering
\begin{tabular}{|cc|ccc|ccc|}
\hline
Method                                             &      & Task 1-a & Task 1-b & Task 1-c & Task 2-a & Task 2-b & Task 2-c \\ \hline \hline
\multicolumn{1}{|c|}{\multirow{3}{*}{Model-based}} & Mean & 2.7982   & 3.1220   & 3.4008   & 2.7232   & 2.6276   & 2.8460   \\
\multicolumn{1}{|c|}{}                             & Std. & 1.5483   & 1.6030   & 1.9731   & 1.6535   & 1.6601   & 1.7303   \\
\multicolumn{1}{|c|}{}                             & Max  & 5.9227   & 6.4958   & 7.1512   & 6.2961   & 6.2279   & 6.4792   \\ \hline \hline
\multicolumn{1}{|c|}{\multirow{3}{*}{Ours}}        & Mean & 0.8825   & 1.0808   & 1.0138   & 1.0189   & 0.9762   & 1.1170   \\
\multicolumn{1}{|c|}{}                             & Std. & 0.7428   & 0.7828   & 0.7851   & 0.8173   & 0.8119   & 0.7561   \\
\multicolumn{1}{|c|}{}                             & Max  & 3.2837   & 3.2535   & 3.8994   & 5.1758   & 8.0854   & 3.4966   \\ \hline
\end{tabular}
\end{table*}


\begin{table*}[]
\captionsetup{justification=centering}
\caption{Experimental results: \textbf{2-norms of endpoint position error (mm) and shape error ($^{\circ}$)} for \texttt{Task 3}}
\small
\label{table:expt3}
\centering
\begin{tabular}{|cc|cc|cc|cc|}
\hline
\multirow{2}{*}{Method}                            &      & \multicolumn{2}{c|}{Task 3-a}           & \multicolumn{2}{c|}{Task 3-b}           & \multicolumn{2}{c|}{Task 3-c}           \\ \cline{3-8} 
                                                   &      & \multicolumn{1}{c|}{Position (mm)} & Shape ($^{\circ}$)   & \multicolumn{1}{c|}{Position (mm)} & Shape ($^{\circ}$)  & \multicolumn{1}{c|}{Position (mm)} & Shape ($^{\circ}$)  \\ \hline \hline
\multicolumn{1}{|c|}{\multirow{3}{*}{Model-based}} & Mean & \multicolumn{1}{c|}{5.1667}   & 7.2742  & \multicolumn{1}{c|}{5.7319}   & 6.7646  & \multicolumn{1}{c|}{6.6612}   & 9.8937  \\
\multicolumn{1}{|c|}{}                             & Std. & \multicolumn{1}{c|}{2.7976}   & 3.9452  & \multicolumn{1}{c|}{3.0444}   & 3.9403  & \multicolumn{1}{c|}{3.0994}   & 6.0184  \\
\multicolumn{1}{|c|}{}                             & Max  & \multicolumn{1}{c|}{11.4878}  & 16.2121 & \multicolumn{1}{c|}{12.8068}  & 19.7125 & \multicolumn{1}{c|}{16.8833}  & 22.4945 \\ \hline \hline
\multicolumn{1}{|c|}{\multirow{3}{*}{Ours}}        & Mean & \multicolumn{1}{c|}{2.4061}   & 3.4782  & \multicolumn{1}{c|}{2.6839}   & 3.6570  & \multicolumn{1}{c|}{2.9202}   & 3.0589  \\
\multicolumn{1}{|c|}{}                             & Std. & \multicolumn{1}{c|}{0.4148}   & 1.4929  & \multicolumn{1}{c|}{0.4267}   & 1.3864  & \multicolumn{1}{c|}{0.5437}   & 1.4099  \\
\multicolumn{1}{|c|}{}                             & Max  & \multicolumn{1}{c|}{3.5322}   & 7.3575  & \multicolumn{1}{c|}{3.8135}   & 7.1389  & \multicolumn{1}{c|}{4.1638}   & 5.8486  \\ \hline
\end{tabular}
\end{table*}

We performed three tasks including \texttt{Task 1}: length sensing in free environments, \texttt{Task 2}: length sensing in unstructured environments, and \texttt{Task 3}: shape sensing with variable sensing length as shown in Fig. \ref{fig:expsetup} (d), (e), and (f), respectively, to evaluate the performances of the proposed sensor and filtering technique.
The experiments of \texttt{Task 2} were conducted in random collisions with a rigid rod as unexpected external disturbances.
In each task, the extensible soft manipulator stretches and compresses in a length of 30 mm under different velocities of (a) 3, (b) 6, and (c) 12 mm/s.
The quantitative metrics, i.e., variable length error, distal endpoint position error, and shape error, were recorded and presented, and each experiment was performed three times.
A video about the experimental results is attached, emphasizing the exterior views of the soft manipulator together with simultaneous shape sensing results using FBG sensors from the model-based method and the proposed approach.

To intuitively compare the performances of the length sensing algorithms for \texttt{Task 1} and \texttt{2}, variable effective lengths of the soft manipulator versus time (iteration) are plotted in Fig. \ref{fig:expt1}.
The length from the ground truth, model-based, and our methods are represented by blue, red, and yellow curves, respectively.
For \texttt{Task 3}, 2-norms of endpoint position errors versus time, i.e., Euclidean distance of the sensing endpoint to the ground truth, are plotted in Fig. \ref{fig:expt3}, where blue and red curves represent the errors using the model-based and our methods, respectively.
As noticed from the results, all sensing lengths using the model-based method have serious deviations with the robot compressing, resulting in obvious position errors of the endpoint.
The sensing lengths of our method are much closer to the ground truth and our position errors are smaller at almost all times and more stable in all experiments regardless of external disturbances and velocities of the robot motion, which preliminarily prove its advancements for variable-length estimation and shape sensing.

For quantitative comparison in each experiment, absolute values of variable lengths for \texttt{Task 1} and \texttt{2} are calculated and presented in Table \ref{table:expt1}, and 2-norms of endpoint errors and shape errors for \texttt{Task 3} are presented in Table \ref{table:expt3}, respectively.
The shape of the soft manipulator for our evaluations is described by bending and twist angles, which are calculated from the constant curvature model using endpoint position of the robot \cite{chen2021tele}.
In \texttt{Task 1} when the soft manipulator changing its length in a free environment as demonstrated in Fig. \ref{fig:expsetup} (d), we can observe that the model-based method outputs large length errors with a maximum mean error as high as 3.4008 mm in experiment (c) under the stretching/compressing velocity of 12 mm/s, which can be explained by the increasing elastic force with the robot compressing makes the variable-length model inaccurate.
The length errors of our method with a maximum mean value of 1.0808 mm can be significantly reduced by approximately three times from that of the model-based method.
When it comes to \texttt{Task 2}, the soft manipulator was deformed in random collisions with a rigid rod considered as unexpected disturbances as shown in Fig. \ref{fig:expsetup} (e).
Notice that the maximum errors listed in the right three columns of Table \ref{table:expt1} represent those caused by the intense and sharp perturbations.
The proposed approach can also outperform with a maximum mean error of 1.1170 mm, reducing that of 2.8460 mm of the model-based method by more than two times.

In \texttt{Task 3}, the soft manipulator deflected under different stretching/compressing velocities as shown in Fig. \ref{fig:expsetup} (f).
As indicated in Table \ref{table:expt3}, we can observe that our method still outputs more accurate results with maximum mean endpoint error of 2.9202 mm and maximum mean shape error of 3.6570$^{\circ}$, while the model-based method outputs more than two times larger errors, which is accordingly 6.6612 mm and 9.8937$^{\circ}$.
The maximum endpoint error of the proposed approach can be maintained within 5 mm (maximum 4.1638 mm) as compared with the model-based one whose error jumped from 11.4878 mm to 16.8833 mm with the increasing velocity.
From qualitative and quantitative results in terms of accuracy and robustness in dynamic conditions, the proposed method outperforms the method using variable-length model, which proves the feasibility and superiority of our algorithm for effective length estimation and shape sensing.

The limitations of the proposed approach are discussed as follows.
Although our method is generic and can be deployed easily for various extensible soft robots, the FBG spatial resolution $\lambda$ \cite{monet2020high} may limit its application to micro-scale robots \cite{chitalia2020towards}.
Besides, the sensing errors could result from the fabrication of the 3-D printed structures for the proposed length sensor and those for attachment of optical markers, which can be solved by improving the printing accuracy.
The unexpected deformation of the FBG fiber within the robot may also defect the results, and we can design additional structures to make the fiber conform to the robot's shape.

\section{Conclusions}
In this paper, we propose a generic variable-length estimation and shape sensing framework for extensible soft manipulators equipped with multi-core FBGs.
An FBG-based length sensor is newly designed to measure the robot's length by matching and locating the FBGs with specific curvature in a rigid channel.
By leveraging the results from our length sensor as measurements, we introduce a novel model-free filtering algorithm to iteratively calibrate a variable-length model and simultaneously estimate the effective length, which can guarantee the length sensing continuity in the temporal domain, and the results can be further utilized for shape sensing.
The accuracy and robustness of the length estimation, as well as shape reconstruction in dynamic conditions, have been evaluated through comprehensive experiments using an extensible soft manipulator both in free space and in unstructured environments, supporting the claim of the feasibility and superiority of our new method.

For our future research, we will advance the performances of the proposed framework for variable-length estimation and shape sensing.
To be more specific, we will improve the fabrication accuracy, select more suitable materials, and optimize the structure parameters of the proposed length sensor.
External sensing units such as eye-in-hand camera can be integrated to provide more efficient data, enabling visual servoing task \cite{wang2020eye} with length control for extensible soft robot to be investigated.

\addtolength{\textheight}{-12cm}   

\bibliographystyle{IEEEtran}
\bibliography{./references}

\end{document}